\documentclass[conference]{IEEEtran}
\IEEEoverridecommandlockouts
\usepackage{cite}
\usepackage{amsmath,amssymb,amsfonts}
\usepackage{algorithmic}
\usepackage{graphicx}
\usepackage{textcomp}
\usepackage{xcolor}
\usepackage{subfig}
\usepackage{dsfont}
\usepackage{multirow}
\usepackage[normalem]{ulem}
\useunder{\uline}{\ul}{}
\def\BibTeX{{\rm B\kern-.05em{\sc i\kern-.025em b}\kern-.08em
    T\kern-.1667em\lower.7ex\hbox{E}\kern-.125emX}}
\begin{document}

\title{Gradient-based Learning in State-based Potential Games for Self-Learning Production Systems}

\author{
\IEEEauthorblockN{1\textsuperscript{st} Steve Yuwono, 2\textsuperscript{nd} Marlon L{\"o}ppenberg, \\ 4\textsuperscript{th} Andreas Schwung}
\IEEEauthorblockA{\textit{Automation Technology and Learning Systems} \\
\textit{South Westphalia University of Applied Sciences}\\
Soest, Germany \\
\{yuwono.steve, loeppenberg.marlon, schwung.andreas\}@fh-swf.de}
\and
\IEEEauthorblockN{3\textsuperscript{rd} Dorothea Schwung}
\IEEEauthorblockA{\textit{Artificial Intelligence and Data Science} \\ \textit{in Automation Technology} \\
\textit{Hochschule D{\"u}sseldorf University of Applied Sciences}\\
D{\"u}sseldorf, Germany \\
dorothea.schwung@hs-duesseldorf.de}
}

\maketitle

\begin{abstract}
In this paper, we introduce novel gradient-based optimization methods for state-based potential games (SbPGs) within self-learning distributed production systems. SbPGs are recognised for their efficacy in enabling self-optimizing distributed multi-agent systems and offer a proven convergence guarantee, which facilitates collaborative player efforts towards global objectives. Our study strives to replace conventional ad-hoc random exploration-based learning in SbPGs with contemporary gradient-based approaches, which aim for faster convergence and smoother exploration dynamics, thereby shortening training duration while upholding the efficacy of SbPGs. Moreover, we propose three distinct variants for estimating the objective function of gradient-based learning, each developed to suit the unique characteristics of the systems under consideration. To validate our methodology, we apply it to a laboratory testbed, namely Bulk Good Laboratory Plant, which represents a smart and flexible distributed multi-agent production system. The incorporation of gradient-based learning in SbPGs reduces training times and achieves more optimal policies than its baseline.

\end{abstract}


\section{Introduction}\label{sec:intro}

The emergence of the Internet of Things (IoT), cyber-physical systems, and artificial intelligence (AI) has transformed production environments, which enhances efficiency and adaptability~\cite{Pivoto2021}. These technologies have revolutionized various sectors by improving productivity, quality, and safety while reducing costs. For instance, IoT sensors enable real-time monitoring of production processes~\cite{Thakare2023}. Similarly, AI-powered algorithms boost decision-making as in~\cite{Chen2021}. In manufacturing, production systems are often characterized by complex control parameters and considered multi-agent systems (MAS)~\cite{Wooldridge2009} with multi-objective optimization challenges~\cite{Branke2008}. Such complexity requires agile and adaptive control mechanisms, which can be realized through the adoption of distributed architectures with decentralizing control systems. Distributed production systems are applicable to a wide range of sectors within the process industry, e.g. food production, chemical plants, pharmaceuticals, and the transportation of grains and coal.

Recent years have witnessed a huge demand for self-learning capabilities within distributed production systems~\cite{Yuwono2023a, Scrimieri2021} through learning from experiences. Advanced methodologies have been developed to meet this demand. Deep learning, particularly through multi-agent reinforcement learning (RL)~\cite{Sutton2018} is the foremost among these. For instance, multi-agent RL for energy-optimal production policies~\cite{Schwung2018}, RL-based scheduler for dynamic manufacturing jobs~\cite{Zhou2020}, and federated RL for distributed energy management~\cite{Lee2020}. Despite such advancements, real-world applications remain limited due to lengthy training times and complex training processes, involving numerous parameters and settings~\cite{Nguyen2020}. Furthermore, the effectiveness of multi-agent RL in collaborative environments is arguable, as agents often operate independently and focus on optimising individual objectives. Consequently, a challenge in MAS is in enabling collaborative behaviour among agents aligned with overarching system objectives.

An alternative self-learning approach gaining traction in distributed learning MAS involves integrating game theoretical (GT) methods, as in~\cite{Cappello2021, Wang2022, Schwung2020}. Among these, state-based potential games (SbPGs) with best response learning~\cite{Schwung2020} have appeared particularly successful, which offer a simpler structure than deep learning and better suitability for real-world applications. Despite their success, there is room for improvement, notably in optimizing the exploration mechanism during policy training to faster convergence while improving performance. In best response learning~\cite{Schwung2020}, players operate random sampling during the exploration phase, which may lead to inefficiencies and lack of exploration direction, which increases training times and slows policy convergence.

Hence, our study aims to introduce a novel method for enhancing SbPGs by integrating gradient-based learning, which effectively guides players' exploration direction, leading to faster and smoother convergence compared to best response learning. We also propose solutions to deal with the challenge of unknown and non-convex utility functions inherent in gradient-based learning by developing multiple estimation function variants. Here are the contributions of our study:
\begin{itemize}
    \item We introduce gradient-based learning within SbPGs for self-learning in distributed production systems, which employs Gradient Ascent with Newton’s First Divided Difference Method.
    \item We present three variants of estimation function to accommodate various characteristics of the objective function, which are the basic method, augmented with momentum, and incorporating polynomial interpolation.
    \item We validate the proposed methods through application in a laboratory testbed and a comparative analysis with best response learning.
\end{itemize}

The paper is structured as follows. In Sec.~\ref{sec:review}, we provide preliminary research. Sec.~\ref{sec:funda} presents the fundamentals of SbPGs and best response learning. Sec.~\ref{sec:method} elaborates on the proposed gradient-based learning within SbPGs. Sec.~\ref{sec:res} presents the results and discussions of our experimentation. Finally, in Sec.~\ref{sec:conc}, we draw conclusions based on our findings.

\section{Preliminary Research}\label{sec:review}

In this section, we explore relevant literature, including GT for distributed optimization and gradient-based optimization.

\subsection{Game Theory for Distributed Optimization}\label{sec:review_1}

In recent years, GT has expanded beyond its origin domain in economics to contain engineering disciplines~\cite{Bauso2016}, which plays a crucial role in analyzing and optimizing complex systems involving multiple decision-makers, such as resource allocation~\cite{Zhang2021}, autonomous driving~\cite{Hang2020}, and cyber security protocols~\cite{Kamhoua2021}. GT has facilitated self-learning within distributed production systems, as shown in SbPGs~\cite{Schwung2020}. Previous studies~\cite{Marden2012, Zazo2016} have spread foundational frameworks for analyzing the convergence of equilibrium points in SbPGs and defining conditions for dynamic and adaptive games, which have further extended for distributed self-learning MAS in~\cite{Schwung2020}. Moreover, we have expanded SbPGs by integrating communication and memory-based learning mechanisms~\cite{Yuwono2022} as well as model-based learning domains~\cite{Yuwono2023a, Yuwono2023b}. However, a limitation remains due to the dependency on best response learning~\cite{Schwung2020}, where random actions are chosen uniformly during exploration. This results in slower and unstable learning behaviours. Hence, our study aims to enhance training efficiency and smoothness through gradient-based learning, while preserving SbPGs as foundational game structures of distributed self-learning.

\subsection{Gradient-based Optimization}\label{sec:review_2}

Gradient ascent~\cite{Lin2020}, a cornerstone of optimization, involves iteratively adjusting parameters to maximize an objective function by following its positive gradient direction, with variants~\cite{Beznosikov2023}, e.g. stochastic and batch gradient ascent. In contrast, gradient descent~\cite{Lin2020} operates oppositely. Various optimization algorithms, like Adam, RMSprop, and Adagrad, provide strategies for efficiently updating model parameters, thereby improving convergence in optimization tasks~\cite{Haji02021}. These also find application in machine learning, which serves as the backbone for training neural networks~\cite{Haji02021}, optimizes support vector machines~\cite{Gieseke2014}, and fits regression models~\cite{Zemel2000}. However, when applied to non-convex problems, challenges arise in locating global optima among multiple local minima/maxima~\cite{Lin2020}. Overcoming these challenges requires strategies like warm restarts~\cite{Loshchilov2016} and annealing schedules~\cite{Pan2020}. Moreover, gradient-based optimization has been theoretically proven to achieve faster and smoother convergence~\cite{Ruder2016}, as demonstrated in applications such as convolutional neural networks~\cite{Dogo2018} and nonlinear optimization methods~\cite{Watanabe2018}, to name a few. In this study, we present gradient-based optimization techniques within SbPGs to achieve faster convergence, improve the balance between exploration and exploitation, and enhance resource utilization compared to the undirected nature of random sampling.

\section{Fundamentals of State-based Potential Games and Best Response Learning}\label{sec:funda}

In this section, we explain the fundamental game structure of SbPGs and its learning algorithm, best response learning.

\subsection{State-based Potential Games}\label{sec:funda_1}

We start with the fundamental principles of SbPGs, followed by their application in optimizing distributed manufacturing processes. SbPGs represent a subset of potential games (PGs)~\cite{Monderer1996}. A PG contains a set of \textit{N} players, denoted individually as $i=1, \ldots, N$. Unlike other games, where each player's utility function $u_i$ is influenced solely by their individual actions $a_i$, in PGs, the utility function $u_i$ in PGs is influenced by the global state of the entire system. An additional component of PGs is a scalar potential function $\phi$, also known as the global objective function. Consequently, the strategic-form game of PGs can be formulated as $\Gamma(\mathcal{N}, A, \{u_i\}, \phi)$, where $A$ represents the set of individual player actions $a_i$.

Furthermore, in SbPGs~\cite{Schwung2020}, the set of states \textit{S} and the state transition process \textit{P} are integrated into the game. To ensure SbPGs hold to the fundamental theorem of PGs, the potential function $\phi: a \times S \rightarrow \mathcal{R}^{c_i}$ must be available within the game structure. Furthermore, each action-state pairing $[a,s] \in A \times S$ must satisfy the following conditions:
\begin{equation}\label{eq:potcondsbpg}
u_i(a_i,s) - u_i({a}^{\prime}_i, {a}_{-i},s) = \phi(a_i,s) - \phi({a}^{\prime}_i, {a}_{-i},s),
\end{equation}
and
\begin{equation}\label{eq:condsbpg}
\phi(a_i,s^{\prime}) \geq \phi(a_i,s),
\end{equation}
for any state $s^{\prime}$ in $P(a,s)$ as well as $\mathcal{R}^{c_i}$ denotes continuous actions.

\begin{figure}[t]
 \centering
 \includegraphics[width=1.00\columnwidth,keepaspectratio]{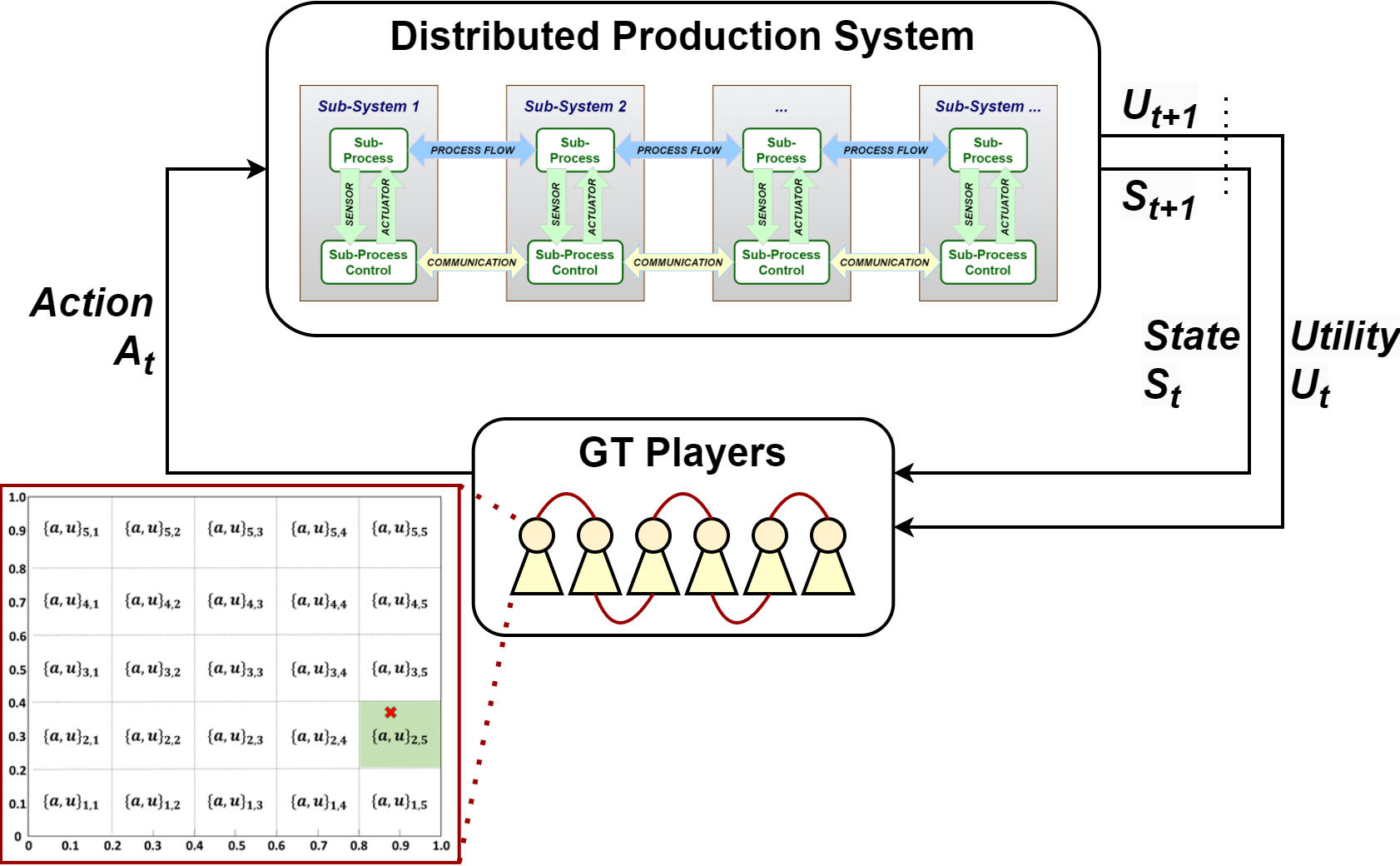}
\caption{An overview of self-learning mechanism in distributed production systems using a game structure of SbPGs.}
\label{fig:mdp}
\end{figure}

In~\cite{Schwung2020}, we demonstrate their application in manufacturing optimization settings and prove the existence of its convergent, as illustrated in Fig.~\ref{fig:mdp}. The distributed system under consideration is characterised and examined using graph theory~\cite{Yamamoto2015}. Further details regarding SbPGs in a production chain of manufacturing graph are presented in~\cite{Schwung2020, Yuwono2022, Yuwono2023a}. The overarching goal is to optimize production systems in a fully distributed manner, avoiding reliance on a centralized instance, which remains consistent throughout this study.

\subsection{Best Response Learning}\label{sec:funda_2}

In~\cite{Schwung2020}, best response learning was developed as an algorithm for training the policy of each player within SbPGs. A performance map is used to record the players' experiences and visualize the distribution of actions across the state space, as shown in Fig.~\ref{fig:mdp}. This map discretizes the state space into support vectors, $l=1,\ldots,L$. Each support vector retains the best-explored and its utility value within the related state. The performance maps are updated, as follows:
\begin{equation}\label{eq:update_maps_1}
u_{i,max}^l(s_i)=u_{i,k}^l(s_i)\text{, if } u_{i,k}^l(s_i) > u_{i,max}^l(s_i),
\end{equation}
\begin{equation}\label{eq:update_maps_2}
a_{i,max}^l(s_i)=a_{i,k}^l(s_i)\text{, if } u_{i,k}^l(s_i) > u_{i,max}^l(s_i),
\end{equation}
in which $u_{i,k}^l$ and $a_{i,k}^l$ represents the utility and action resulted by player $i$ in the current iteration $k$, associated with the $l$-th support vector.

During exploration, actions $a_{i,k}^l(s_i)$ are computed based on random uniform sampling, which results in utilities $u_{i,k}^l(s_i)$. This process allows agents to gather experiences and store the best-explored actions in the maps. Meanwhile, during exploitation, each player $i$ selects an action $a_{i,t+1}$ by globally interpolating the performance map according to its current state. The global interpolation rule is managed, as follows:
\begin{equation} 
D_i^{S^0S^k} = \sqrt{{(S_x^0 - S_x^k)}^2 + {(S_y^0 - S_y^k)}^2},
\end{equation}
\begin{equation} 
w_i^{\overline{S^0S^k}} = \dfrac{1}{{(D_i^{S^0S^k})}^2 + \gamma},
\end{equation}
\begin{equation} 
a_{i,t+1} = \sum_{k} \dfrac{w_i^{\overline{S^0S^k}}}{\sum_{m}w_i^{S^0S^m}} \cdot a_i^k,
\end{equation}
in which \textit{x} and \textit{y} denote the prior and subsequent states, \textit{n} represents the player identifier, and \textit{$D_i^{S^0S^k}$} indicates the distance from a support vector to the current state combination, while \textit{$w_i^{\overline{S^0S^k}}$} calculates the weighted value based on this distance. Here, \textit{$S^0$} represents the current state combination, \textit{$S^k$} denotes the state combination of the centre \textit{k}, and \textit{$\gamma$} represents the smoothing parameter for the performance map.

\section{Gradient-based Learning in State-based Potential Games}\label{sec:method}

\begin{figure}[t]
 \centering
 \subfloat[Best response learning]{\includegraphics[width=0.46\columnwidth,keepaspectratio]{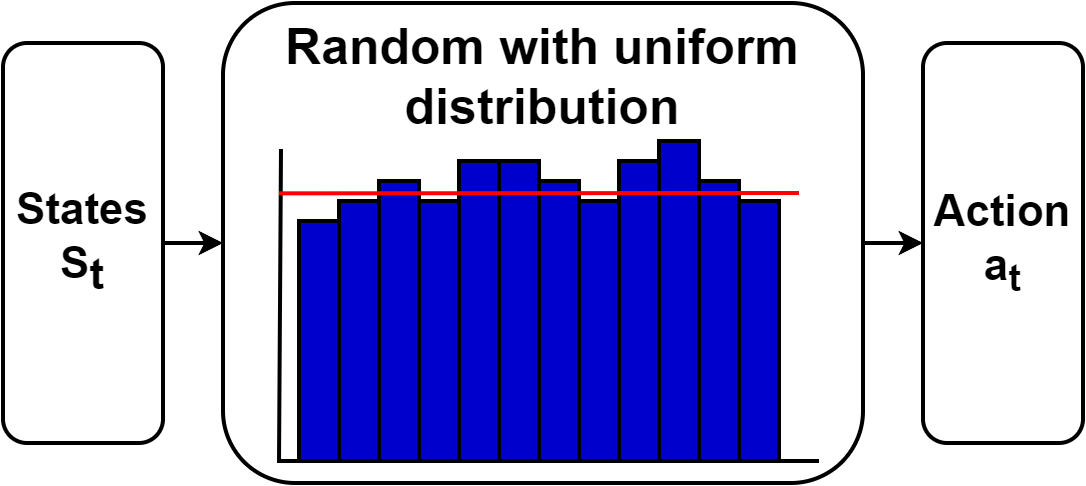}} 
  \qquad 
 \subfloat[Gradient-based learning]{\includegraphics[width=0.46\columnwidth,keepaspectratio]{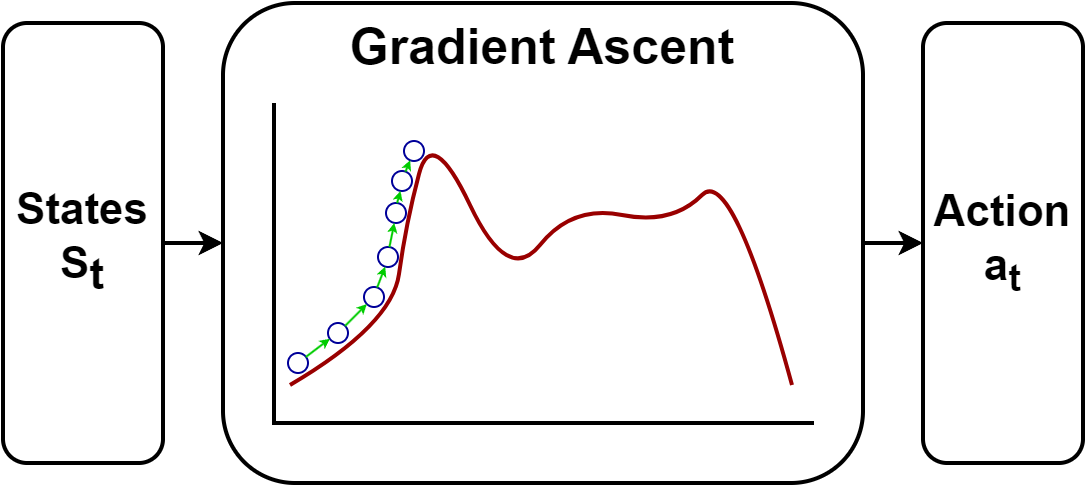}}
\caption{Learning methods during exploration in SbPGs.}
\label{fig:br_grad}
\end{figure}

\begin{figure}[t]
 \centering
 \includegraphics[width=0.90\columnwidth,keepaspectratio]{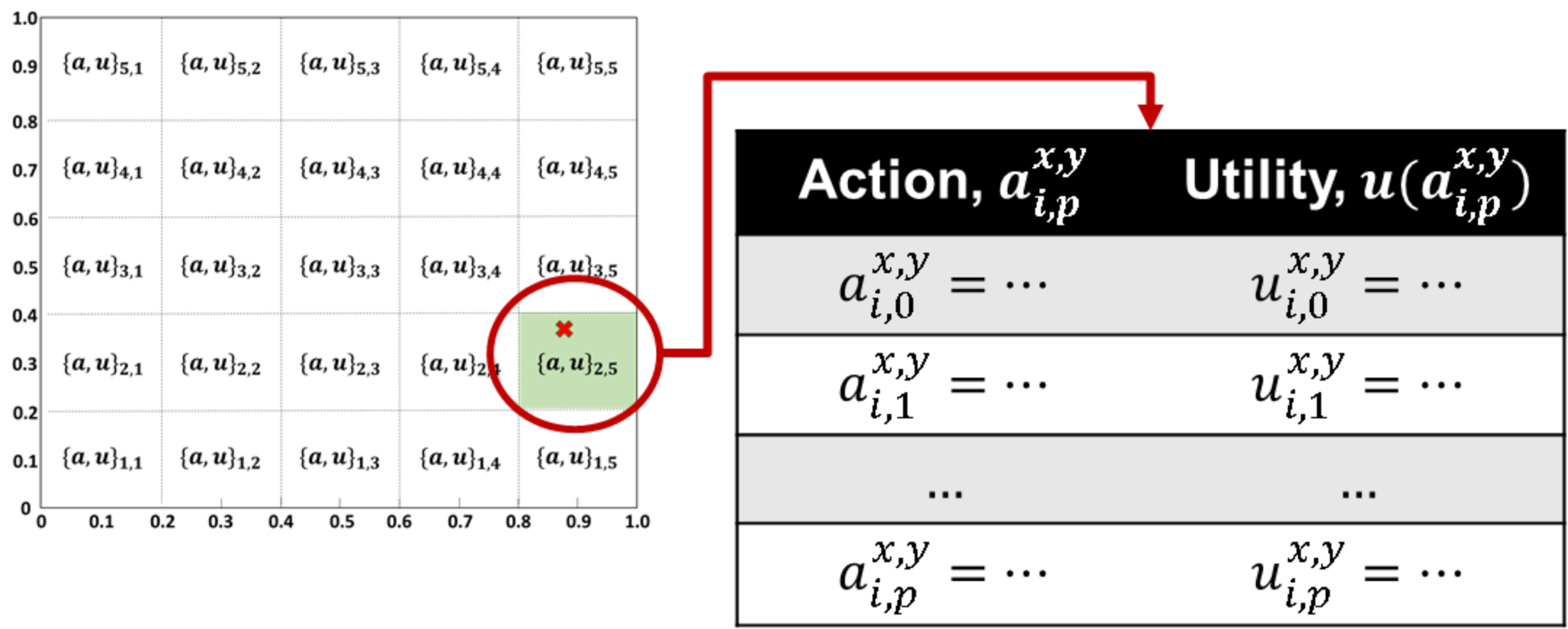}
\caption{An example of a 5 × 5 performance map with a 2D-state space in SbPGs with gradient-based learning.}
\label{fig:updated_maps}
\end{figure}

This study introduces novel gradient-based methods to enhance exploration in SbPGs, which replaces random sampling in best response learning. Fig.~\ref{fig:br_grad} highlights the contrast between best response and gradient-based learning. Initially, the performance map structure is modified to accommodate the new approach. Instead of storing only the best-explored action and its corresponding utility value in each state combination, we now stack selected actions and their utilities for each data point across various state combinations, as shown in Fig.~\ref{fig:updated_maps}. Here, $x$ and $y$ denote the position in the 2D grid (but not limited to 2D), $i$ represents the index of the relevant player, and $p$ indicates the actual size of the selected actions. Additionally, this leads to different updated rules for the performance maps.

In the gradient-based learning approach, actions are considered as weights to be optimized with respect to the utility function, which serves as the objective function. The adjustment of actions is guided by the gradient of the utility function $\nabla J$. To promote stable exploration, Ornstein-Uhlenbeck (OU) noise~\cite{Uhlenbeck1930} is optionally incorporated, which introduces temporally correlated noise that prevents rapid and erratic changes in actions. OU noise also helps in overcoming local optima and exploring a wider range of states, which leads to more robust learning, particularly in continuous action spaces. In this study, OU noise can be optionally activated during exploration but is deactivated during exploitation. The impact of OU noise on our proposed method is further analyzed in our experiments. Therefore, actions during exploration are computed as follows:
\begin{equation}\label{eq:action_grad}
    a_{i,p+1}^{x,y} = a_{i,p}^{x,y} + \alpha \cdot \nabla J + \gamma_{ou},
\end{equation}
where $a_{i,p}^{x,y}$ represents the previous action in the $x,y$ state combination, $a_{i,p+1}^{x,y}$ denotes the new action, $\gamma_{ou}$ signifies the OU noise output, and $\alpha$ denotes the learning rate. Initially, each action in each data point is set to zero, as follows:
\begin{equation}\label{eq:init_weight}
    a_{i,0}^{x,y} = u_{i,0}^{x,y} = 0.
\end{equation}
The gradient-based learning automatically cancels the update rules in Eq.~\eqref{eq:update_maps_1} and~\eqref{eq:update_maps_2}. By cancelling those, gradient-based learning achieves faster convergence and smoother exploration dynamics than best-response learning. However, the learning rate $\alpha$ must be properly defined.

The challenge is in dealing with non-convex or unknown mathematical formulas of utility functions for each player within SbPGs. Players often face unknown mathematical formulas of utility functions as these functions are integral to the system and can be challenging to derive. As in Fig.~\ref{fig:mdp}, the players interact with the system without knowing their utility function. Instead, they learn and optimize their policies based on the utility values (not functions) associated with selected actions in particular states. Hence, players must internally estimate the utility function to provide the gradient-based learner with directional learning signals. In response to this challenge, we propose three different estimation variants for different objective functions, systems, and complexities. These variants utilize Newton's first divided difference method~\cite{Epperson2013}.

In the following subsections, we discuss the three proposed variants, including the basic estimation method, augmented with momentum, and incorporating polynomial interpolation. These three variants operate characteristic estimation function structures, each offering unique advantages. The basic estimation method calculates gradients directly from the current iteration's utility. Momentum enhances this process by integrating a portion of the previous iteration's gradient to smooth out fluctuations and accelerate convergence. Incorporating polynomial interpolation takes a step further by utilizing a polynomial curve to represent historical gradient data, which potentially offers a more subtle and flexible approach to gradient estimation. Additionally, we introduce a kick-off method aimed at accelerating the training process.

\subsection{Gradient Ascent with Newton's First Divided Difference Method}\label{sec:method_1}

The first variant, basic gradient ascent with Newton's first divided difference method, offers advantages such as reduced memory usage, suitability for estimations of convex utility functions, and faster convergence. However, it is easy to get trapped in local minima and struggle with handling sensitivity to noise or fluctuations in the objective function. The estimation of the utility function significantly influences the calculation of the gradient $\nabla J$, as depicted below:
\begin{equation}\label{eq:ga_1_1}
    \nabla J=\frac{u(a_{i,p}^{x,y})-u(a_{i,p-1}^{x,y})}{a_{i,p}^{x,y}-a_{i,p-1}^{x,y}}\text{, if }  a_{i,p}^{x,y} \neq a_{i,p-1}^{x,y}
\end{equation}
\begin{equation}\label{eq:ga_1_2}
    \nabla J=u(a_{i,p}^{x,y})-u(a_{i,p-1}^{x,y})\text{, if }  a_{i,p}^{x,y} = a_{i,p-1}^{x,y}
\end{equation}

\subsection{Gradient Ascent with Newton's First Divided Difference Method and Momentum}\label{sec:method_2}

The second variant builds upon the previous approach by incorporating momentum, with expected benefits including suitability for estimations of non-convex utility functions with less memory usage, comfort of oscillations, faster convergence by smoothing the optimization trajectory and allowing rapid adaptation to the gradient landscape. However, drawbacks include the risk of overshooting optimal solutions and sensitivity to hyperparameters. In this variant, Eq.~\eqref{eq:action_grad} remains applicable, but the gradient of the utility function $\nabla J$ is now denoted as $\nabla J_p$:
\begin{equation}\label{eq:action_grad_mom}
    a_{i,p+1}^{x,y} = a_{i,p}^{x,y} + \alpha \cdot \nabla J_p + \gamma_{ou},
\end{equation}
where $\nabla J_p$ is calculated based on momentum:
\begin{equation}\label{eq:grad_mom}
    \nabla J_p = \beta \cdot \nabla J_{p-1} + (1-\beta) \cdot \nabla J
\end{equation}
with $\beta$ representing the momentum factor weighting. The computation of $\nabla J$ remains consistent with Eq.~\eqref{eq:ga_1_1} and~\eqref{eq:ga_1_2}.

\subsection{Gradient Ascent with Newton's First Divided Difference Method of Polynomial Interpolation}\label{sec:method_3}

The third variant incorporates polynomial interpolation, which results in a more precise approximation of the objective function's landscape and facilitates smoother and more efficient exploration of the optimization space. However, this enhancement introduces increased computational complexity and memory requirements, along with potential challenges in selecting appropriate polynomial degrees or coefficients. In this variant, we do not utilize momentum, which allows us to use Eq.~\eqref{eq:action_grad}. The objective function is estimated using interpolating polynomial forms, which derives the following equation to compute the gradient of the utility function $\nabla J$:
\begin{equation}
\begin{split}
    \nabla J & = u [ a_{i,0}^{x,y},a_{i,1}^{x,y},\dots,a_{i,p}^{x,y} ] \\ 
    & = \frac{u [ a_{i,1}^{x,y},a_{i,1}^{x,y},\dots,a_{i,p}^{x,y} ]-u [ a_{i,0}^{x,y},a_{i,1}^{x,y},\dots,a_{i,p-1}^{x,y} ]}{a_{i,p}^{x,y} - a_{i,0}^{x,y}},
\end{split}
\end{equation}
where $u [ a_{i,0}^{x,y},a_{i,1}^{x,y},\dots,a_{i,p}^{x,y} ]$ represents the recursive generation of the divided differences, with the bracket notation introduced to differentiate these differences. This notation starts from $u[a_{i,0}^{x,y}] = u(a_{i,0}^{x,y})$.

\subsection{Kick-Off with Random Exploration}\label{sec:method_4}

\begin{figure}[t]
 \centering
 \includegraphics[width=0.9\columnwidth,keepaspectratio]{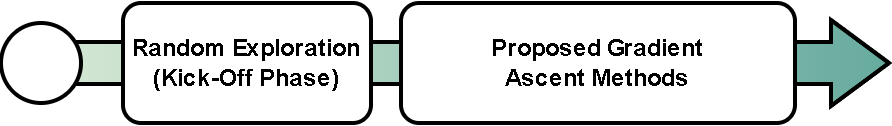}
\caption{An illustration of kick-off procedure in SbPGs with gradient-based learning.}
\label{fig:kickoff}
\end{figure}

Similar to gradient-based optimization in neural networks, often initialized with random weights, which has been proven effective~\cite{Chen2019}, we propose introducing a kick-off start for the proposed gradient-based learning in SbPGs, as pictured in Fig.~\ref{fig:kickoff}. The concept involves beginning with a period of random exploration, similar to best response learning, before transitioning to the proposed gradient ascent method. This kick-off mechanism causes Eq.~\eqref{eq:init_weight} invalid.

\section{Results and Discussions}\label{sec:res}

In this section, we introduce the laboratory testbed used for evaluating the proposed methods. Then, we present the results of benchmark tests conducted on the testbed. Finally, we provide the results and discussions of the proposed methods.

\subsection{Bulk Good Laboratory Plant}\label{sec:res_1}

\begin{figure*}[t]
 \centering
 \includegraphics[width=2.00\columnwidth,keepaspectratio]{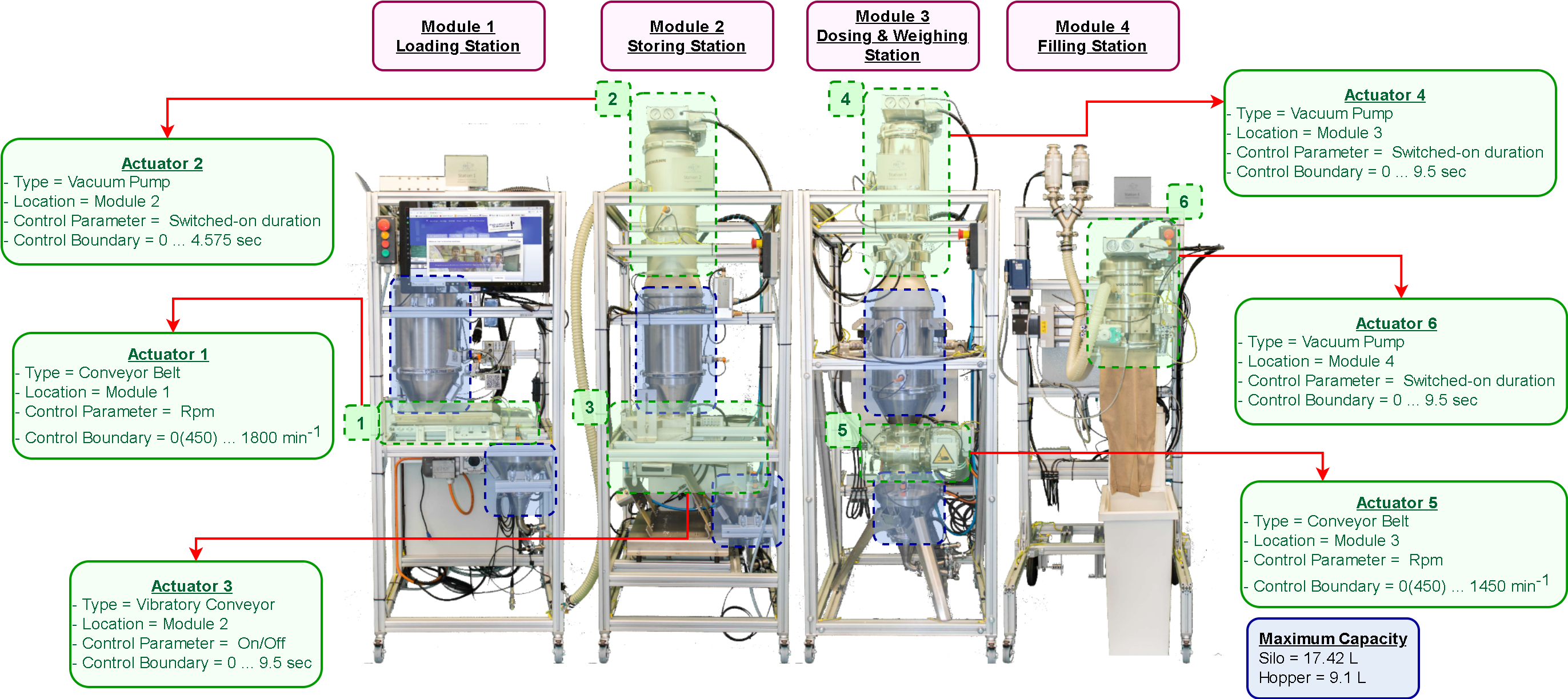}
\caption{Bulk Good Laboratory Plant.}
\label{fig:bglp}
\end{figure*}

The Bulk Good Laboratory Plant (BGLP)~\cite{Schwung2020, Schwung2022} represents a modular distributed production system characterized by its intelligence, flexibility, and plug-and-play functionality~\cite{Yuwono2023a}. The basic setup comprises four main modules, which are loading, storing, weighing, and filling stations. They are arranged sequentially and equipped with different actuators, as described in Fig.~\ref{fig:bglp}. Each station features a silo and hopper with maximum capacities of 17.42L and 9.1L, respectively, where the actuators facilitate the transfer of materials between buffers. Additionally, a PLC-based Siemens ET200SP control system is integrated into each station, which facilitates communication via Profinet. Furthermore, a group of sensors is installed to monitor the current level of each buffer, which are later used as states in SbPGs. The BGLP includes a feature that utilizes the You Only Look Once (YOLO) v8 model to detect foreign objects in the system and activate an ejection system when they are identified~\cite{Loppenberg2024}.

In SbPGs, each actuator is considered as one player. The utility function $U_i$ formulated in~\cite{Schwung2020, Yuwono2022, Yuwono2023a} is designed to combine the multiple objectives of the system, which remains in this study. The utility function $U_i$ is formulated, as follows:

\small
\begin{equation}
    U_i = \frac{1}{1+\alpha_{l}L_{p}^{i}}+\mathds{1}_{i\neq N}\frac{1}{1+\alpha_{l}L_{i}^{n}} + \mathds{1}_{i=N}\frac{1}{1-\alpha_{d}V_{D}}  +\frac{1}{1+\alpha_{p}P^{i}},
\end{equation}
\normalsize
in which $P^{i}$ is the power consumption of actuator $i$, $\mathds{1}_{i=N}$ denotes the identity function, where $i=N$ shows the last player in the sequence, $V_{D}$ represents the fulfilled production demand, and $\alpha_l$, $\alpha_d$, and $\alpha_p$ denote weighting parameters for each objective. Constraints $L_{s}^{i}$ and $L_{p}^{i}$ are computed to prevent overflow and bottleneck based on the upper and lower limits of the corresponding buffer's level. To be noted, the utility function $U_i$ is integral to the system, see Fig.~\ref{fig:mdp}. Therefore, the gradient-based learning player cannot access its utility function and must resort to estimation methods outlined in Sec.~\ref{sec:method}.

In our experiments, we implement a continuous production process with a production demand set at 0.110L/s. Each method undergoes a maximum of 20 training episodes and 1 testing episode, with each episode lasting 10,000 seconds. Each player computes a new action every 10 seconds. Additionally, we discretize the state space into 40. Furthermore, all parameters related to the algorithms and learning approaches pass automated tuning using Hyperopt~\cite{Bergstra2013}. Then, the simulation model of the BGLP is publicly accessible through both the MLPro~\cite{Arend2022} and MLPro-MPPS~\cite{Yuwono2023c} frameworks. In this study, we conduct policy training within the simulation model due to considerations of safety, cost, and time constraints. However, the policy trained in the simulation can be deployed in the real system within our laboratory.

\subsection{Benchmark: SbPGs with Best Response Learning}\label{sec:res_2}

We apply the SbPGs approach with best response learning~\cite{Schwung2020} as our benchmark for comparison with the proposed gradient-based learning method. Through experimentation, we achieve the best results when training the players over 20 episodes. In the testing episode, we observe complete avoidance of overflow, with an average power consumption of 0.475885 kW/s, accomplishment of the production demand of 0.110L/s, and an average potential value of 12.905199. These outcomes suggest that the resulting policies derived from SbPGs with best response learning are nearly optimal. However, the training duration remains lengthy and the exploration behaviour remains uncontrollable.

\subsection{Results on Gradient-based Learning in SbPGs}\label{sec:res_3}

In this subsection, we present the results of the gradient-based learning in SbPGs for all three variants, considering both with and without kick-off for each variant. The number of kick-off episodes is denoted as $\theta_{k}$. Table~\ref{tab:res} summarizes the testing results of all variants and provides comparisons between the best response and gradient-based learning in the BGLP.
\begin{table}[t]
\renewcommand{\arraystretch}{1.2}
\caption{Results and comparisons between best response and gradient-based learning in the BGLP.}
\label{tab:res}
	\centering
\begin{tabular}{|cccccc|}
\hline
\multicolumn{1}{|c|}{\textbf{\begin{tabular}[c]{@{}c@{}}Method\\ {[}$E_{ko}${]}\end{tabular}}} & \multicolumn{1}{c|}{\textbf{\begin{tabular}[c]{@{}c@{}}Training\\ Time {[}s{]}\end{tabular}}} & \multicolumn{1}{c|}{\textbf{\begin{tabular}[c]{@{}c@{}}Overflow\\ {[}L/s{]}\end{tabular}}} & \multicolumn{1}{c|}{\textbf{\begin{tabular}[c]{@{}c@{}}Power\\ {[}kW/s{]}\end{tabular}}} & \multicolumn{1}{c|}{\textbf{\begin{tabular}[c]{@{}c@{}}Demand\\ {[}L/s{]}\end{tabular}}} & \textbf{\begin{tabular}[c]{@{}c@{}}Potential\\ {[}-{]}\end{tabular}} \\ \hline\hline
\multicolumn{6}{|c|}{{\ul \textbf{Benchmark: Best Response Learning}}} \\ \hline
\multicolumn{1}{|c|}{-} & \multicolumn{1}{c|}{200,000} & \multicolumn{1}{c|}{0.0000} & \multicolumn{1}{c|}{0.4759} & \multicolumn{1}{c|}{0.0000} & 12.9052 \\ \hline\hline
\multicolumn{6}{|c|}{{\ul \textbf{Gradient-based Learning}}} \\ \hline
\multicolumn{1}{|c|}{\begin{tabular}[c]{@{}c@{}}1 {[}$\theta_k=0${]}\end{tabular}} & \multicolumn{1}{c|}{200,000} & \multicolumn{1}{c|}{0.0000} & \multicolumn{1}{c|}{0.4374} & \multicolumn{1}{c|}{0.0000} & 12.4245 \\ \hline
\multicolumn{1}{|c|}{\begin{tabular}[c]{@{}c@{}}1 {[}$\theta_k=3${]}\end{tabular}} & \multicolumn{1}{c|}{120,000} & \multicolumn{1}{c|}{0.0000} & \multicolumn{1}{c|}{0.4317} & \multicolumn{1}{c|}{0.0000} & 12.4275 \\ \hline\hline
\multicolumn{1}{|c|}{\begin{tabular}[c]{@{}c@{}}2 {[}$\theta_k=0${]}\end{tabular}} & \multicolumn{1}{c|}{200,000} & \multicolumn{1}{c|}{0.0000} & \multicolumn{1}{c|}{0.4415} & \multicolumn{1}{c|}{0.0000} & 13.0255 \\ \hline
\multicolumn{1}{|c|}{\begin{tabular}[c]{@{}c@{}}2 {[}$\theta_k=4${]}\end{tabular}} & \multicolumn{1}{c|}{120,000} & \multicolumn{1}{c|}{0.0000} & \multicolumn{1}{c|}{0.4477} & \multicolumn{1}{c|}{0.0000} & 12.4286 \\ \hline\hline
\multicolumn{1}{|c|}{\begin{tabular}[c]{@{}c@{}}3 {[}$\theta_k=0${]}\end{tabular}} & \multicolumn{1}{c|}{180,000} & \multicolumn{1}{c|}{0.0000} & \multicolumn{1}{c|}{0.4495} & \multicolumn{1}{c|}{0.0000} & 12.4422 \\ \hline
\multicolumn{1}{|c|}{\begin{tabular}[c]{@{}c@{}}3 {[}$\theta_k=1${]}\end{tabular}} & \multicolumn{1}{c|}{110,000} & \multicolumn{1}{c|}{0.0000} & \multicolumn{1}{c|}{0.4442} & \multicolumn{1}{c|}{0.0000} & 12.3847 \\ \hline
\end{tabular}
\end{table}

In the first variant without kick-off, hyperparameter tuning determined that the optimal parameter combination includes $\alpha$ set to 1.0 and the OU noise ranging approximately within $\pm 0.3$. Conversely, with kick-off, $\alpha$ remains at 1.0, OU noise is deactivated, and $\theta_{k}$ is set to 3. The testing results indicate a reduction in power consumption by approximately 9\% for both approaches compared to the benchmark, which achieves complete overflow avoidance and fulfilling production demand. Additionally, kick-off episodes contribute to a 40\% reduction in training time compared to the benchmark.

In the second variant, both with and without kick-off episodes involve deactivation of the OU noise, with $\alpha$ set to 0.5 and $\beta$ to 0.4. In the variant with kick-off, $\theta_{k}$ is defined as 4. Test results demonstrate a similar trend to the first variant. However, the second variant without kick-off achieves the highest potential value among all variants and the benchmark.

In the third variant, $\alpha$ is set to 0.5, and OU noise is deactivated for both with and without kick-off. In the variant with kick-off, $\theta_{k}$ is set to 1. Testing results show similarities to the previous two variants with shorter training cycles. However, there is an increase in computational time due to the polynomial behaviour.

During the experiments, we investigate the influence of OU noise in our proposed methods. It appears that OU noise could benefit the first variant since it does not consider historical gradient data, which makes it prone to getting stuck in local optima and facing challenges in exploring diverse states. Contrarily, the impact of OU noise on the second and third variants is minimal. As a result, we opt to deactivate OU noise in both variants to simplify parameter selection.

\begin{figure}[t]
 \centering
 \includegraphics[width=1.00\columnwidth,keepaspectratio]{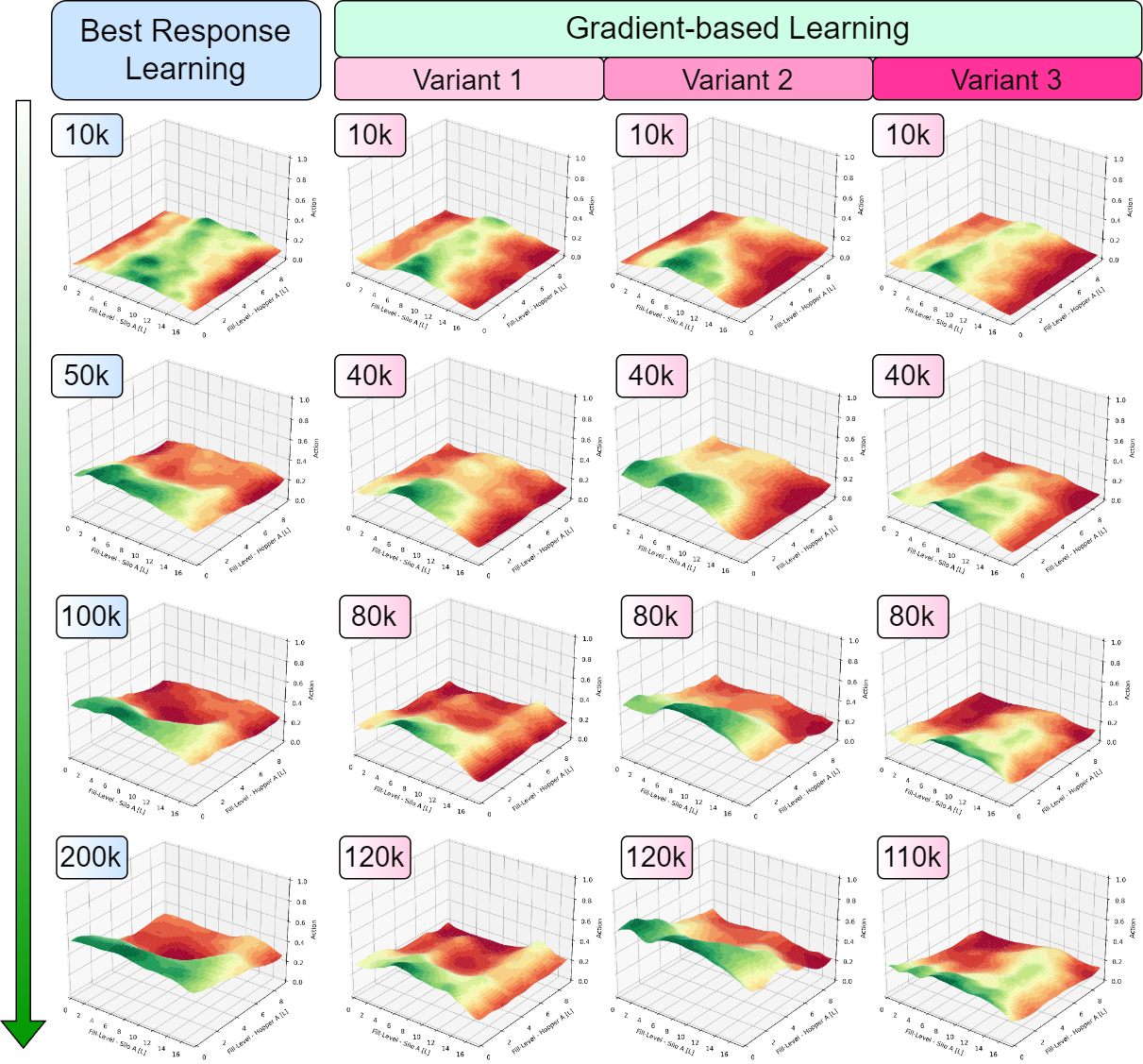}
\caption{A comparison of the resulted actions' performance maps of player 1 (the conveyor belt in Module 1) in the BGLP between best response learning and gradient-based learning.}
\label{fig:compare}
\end{figure}

In Fig.~\ref{fig:compare}, a comparison is defined between the resulting actions' performance maps of player 1 (the conveyor belt in Module 1) in the BGLP. The progress of the performance maps highlights two notable learning characteristics of gradient-based learning compared to best response learning, such as (a) initially slower progression but then followed by faster convergence and (b) a smoother exploration process guided by the gradient-based learners.

\section{Conclusions}\label{sec:conc}

We introduce gradient-based learning methods in SbPGs as a replacement for the ad-hoc random sampling approach used in best response learning during the training of players' policies. These gradient-based learning methods are well-suited for self-learning distributed production systems. We propose three distinct variants, each offering the option of starting with or without a kick-off to establish initial weight values. Afterwards, we apply these variants to a laboratory testbed and conduct comparisons with the benchmark, which is best response learning. Our testing results reveal a significant reduction in power consumption, reaching nearly 10\%, and indicate that one of the variants can even enhance the potential value compared to the benchmark. Additionally, the inclusion of kick-off episodes demonstrates a substantial reduction in training time, up to 45\% compared to the benchmark. These findings highlight the effectiveness and impact of gradient-based learning in SbPGs.

In future works, our objective is to extend the integration of gradient-based learning to model-based SbPGs. Additionally, we plan to explore its integration into other game structures, such as Stackelberg games.

\bibliographystyle{IEEEtran}
\bibliography{reference}

\begin{thebibliography}{10}
\providecommand{\url}[1]{#1}
\csname url@samestyle\endcsname
\providecommand{\newblock}{\relax}
\providecommand{\bibinfo}[2]{#2}
\providecommand{\BIBentrySTDinterwordspacing}{\spaceskip=0pt\relax}
\providecommand{\BIBentryALTinterwordstretchfactor}{4}
\providecommand{\BIBentryALTinterwordspacing}{\spaceskip=\fontdimen2\font plus
\BIBentryALTinterwordstretchfactor\fontdimen3\font minus
  \fontdimen4\font\relax}
\providecommand{\BIBforeignlanguage}[2]{{%
\expandafter\ifx\csname l@#1\endcsname\relax
\typeout{** WARNING: IEEEtran.bst: No hyphenation pattern has been}%
\typeout{** loaded for the language `#1'. Using the pattern for}%
\typeout{** the default language instead.}%
\else
\language=\csname l@#1\endcsname
\fi
#2}}
\providecommand{\BIBdecl}{\relax}
\BIBdecl

\bibitem{Pivoto2021}
D.~G. Pivoto, L.~F. de~Almeida, R.~da~Rosa~Righi, J.~J. Rodrigues, A.~B. Lugli,
  and A.~M. Alberti, ``Cyber-physical systems architectures for industrial
  internet of things applications in industry 4.0: A literature review,''
  \emph{Journal of manufacturing systems}, vol.~58, pp. 176--192, 2021.

\bibitem{Thakare2023}
Y.~N. Thakare, A.~M. Karandikar, V.~Butram, A.~Rokade, U.~A. Wankhade, and
  S.~Honade, ``Iot-enabled environmental intelligence: A smart monitoring
  system,'' in \emph{2023 3rd International Conference on Innovative Mechanisms
  for Industry Applications (ICIMIA)}.\hskip 1em plus 0.5em minus 0.4em\relax
  IEEE, 2023, pp. 37--42.

\bibitem{Chen2021}
W.~Chen, X.~Qiu, T.~Cai, H.-N. Dai, Z.~Zheng, and Y.~Zhang, ``Deep
  reinforcement learning for internet of things: A comprehensive survey,''
  \emph{IEEE Communications Surveys \& Tutorials}, vol.~23, no.~3, pp.
  1659--1692, 2021.

\bibitem{Wooldridge2009}
M.~Wooldridge, \emph{An introduction to multiagent systems}.\hskip 1em plus
  0.5em minus 0.4em\relax John wiley \& sons, 2009.

\bibitem{Branke2008}
J.~Branke, \emph{Multiobjective optimization: Interactive and evolutionary
  approaches}.\hskip 1em plus 0.5em minus 0.4em\relax Springer Science \&
  Business Media, 2008, vol. 5252.

\bibitem{Yuwono2023a}
S.~Yuwono and A.~Schwung, ``Model-based learning on state-based potential games
  for distributed self-optimization of manufacturing systems,'' \emph{Journal
  of Manufacturing Systems}, vol.~71, pp. 474--493, 2023.

\bibitem{Scrimieri2021}
D.~Scrimieri, S.~M. Afazov, and S.~M. Ratchev, ``Design of a self-learning
  multi-agent framework for the adaptation of modular production systems,''
  \emph{The International Journal of Advanced Manufacturing Technology}, vol.
  115, no.~5, pp. 1745--1761, 2021.

\bibitem{Sutton2018}
R.~S. Sutton and A.~G. Barto, \emph{Reinforcement learning: An
  introduction}.\hskip 1em plus 0.5em minus 0.4em\relax MIT press, 2018.

\bibitem{Schwung2018}
D.~Schwung, J.~N. Reimann, A.~Schwung, and S.~X. Ding, ``Self learning in
  flexible manufacturing units: A reinforcement learning approach,'' in
  \emph{2018 International Conference on Intelligent Systems (IS)}.\hskip 1em
  plus 0.5em minus 0.4em\relax IEEE, 2018, pp. 31--38.

\bibitem{Zhou2020}
T.~Zhou, D.~Tang, H.~Zhu, and L.~Wang, ``Reinforcement learning with composite
  rewards for production scheduling in a smart factory,'' \emph{IEEE Access},
  vol.~9, pp. 752--766, 2020.

\bibitem{Lee2020}
S.~Lee and D.-H. Choi, ``Federated reinforcement learning for energy management
  of multiple smart homes with distributed energy resources,'' \emph{IEEE
  Transactions on Industrial Informatics}, vol.~18, no.~1, pp. 488--497, 2020.

\bibitem{Nguyen2020}
T.~T. Nguyen, N.~D. Nguyen, and S.~Nahavandi, ``Deep reinforcement learning for
  multiagent systems: A review of challenges, solutions, and applications,''
  \emph{IEEE transactions on cybernetics}, vol.~50, no.~9, pp. 3826--3839,
  2020.

\bibitem{Cappello2021}
D.~Cappello and T.~Mylvaganam, ``Distributed differential games for control of
  multi-agent systems,'' \emph{IEEE Transactions on Control of Network
  Systems}, vol.~9, no.~2, pp. 635--646, 2021.

\bibitem{Wang2022}
J.~Wang, Y.~Hong, J.~Wang, J.~Xu, Y.~Tang, Q.-L. Han, and J.~Kurths,
  ``Cooperative and competitive multi-agent systems: From optimization to
  games,'' \emph{IEEE/CAA Journal of Automatica Sinica}, vol.~9, no.~5, pp.
  763--783, 2022.

\bibitem{Schwung2020}
D.~Schwung, A.~Schwung, and S.~X. Ding, ``Distributed self-optimization of
  modular production units: A state-based potential game approach,'' \emph{IEEE
  Transactions on Cybernetics}, vol.~52, no.~4, pp. 2174--2185, 2020.

\bibitem{Bauso2016}
D.~Bauso, \emph{Game theory with engineering applications}.\hskip 1em plus
  0.5em minus 0.4em\relax SIAM, 2016.

\bibitem{Zhang2021}
D.-g. Zhang, C.~Chen, Y.-y. Cui, and T.~Zhang, ``New method of energy efficient
  subcarrier allocation based on evolutionary game theory,'' \emph{Mobile
  Networks and Applications}, vol.~26, pp. 523--536, 2021.

\bibitem{Hang2020}
P.~Hang, C.~Lv, Y.~Xing, C.~Huang, and Z.~Hu, ``Human-like decision making for
  autonomous driving: A noncooperative game theoretic approach,'' \emph{IEEE
  Transactions on Intelligent Transportation Systems}, vol.~22, no.~4, pp.
  2076--2087, 2020.

\bibitem{Kamhoua2021}
C.~A. Kamhoua, C.~D. Kiekintveld, F.~Fang, and Q.~Zhu, \emph{Game theory and
  machine learning for cyber security}.\hskip 1em plus 0.5em minus 0.4em\relax
  John Wiley \& Sons, 2021.

\bibitem{Marden2012}
J.~R. Marden, ``State based potential games,'' \emph{Automatica}, vol.~48,
  no.~12, pp. 3075--3088, 2012.

\bibitem{Zazo2016}
S.~Zazo, S.~V. Macua, M.~S{\'a}nchez-Fern{\'a}ndez, and J.~Zazo, ``Dynamic
  potential games with constraints: Fundamentals and applications in
  communications,'' \emph{IEEE Transactions on Signal Processing}, vol.~64,
  no.~14, pp. 3806--3821, 2016.

\bibitem{Yuwono2022}
S.~Yuwono, A.~Schwung, and D.~Schwung, ``The impact of communication and memory
  in state-based potential game-based distributed optimization,'' in \emph{2022
  IEEE 20th International Conference on Industrial Informatics (INDIN)}.\hskip
  1em plus 0.5em minus 0.4em\relax IEEE, 2022, pp. 335--340.

\bibitem{Yuwono2023b}
S.~Yuwono and A.~Schwung, ``A model-based deep learning approach for
  self-learning in smart production systems,'' in \emph{2023 IEEE 28th
  International Conference on Emerging Technologies and Factory Automation
  (ETFA)}.\hskip 1em plus 0.5em minus 0.4em\relax IEEE, 2023, pp. 1--8.

\bibitem{Lin2020}
T.~Lin, C.~Jin, and M.~Jordan, ``On gradient descent ascent for
  nonconvex-concave minimax problems,'' in \emph{International Conference on
  Machine Learning}.\hskip 1em plus 0.5em minus 0.4em\relax PMLR, 2020, pp.
  6083--6093.

\bibitem{Beznosikov2023}
A.~Beznosikov, E.~Gorbunov, H.~Berard, and N.~Loizou, ``Stochastic gradient
  descent-ascent: Unified theory and new efficient methods,'' in
  \emph{International Conference on Artificial Intelligence and
  Statistics}.\hskip 1em plus 0.5em minus 0.4em\relax PMLR, 2023, pp. 172--235.

\bibitem{Haji02021}
S.~H. Haji and A.~M. Abdulazeez, ``Comparison of optimization techniques based
  on gradient descent algorithm: A review,'' \emph{PalArch's Journal of
  Archaeology of Egypt/Egyptology}, vol.~18, no.~4, pp. 2715--2743, 2021.

\bibitem{Gieseke2014}
F.~Gieseke, A.~Airola, T.~Pahikkala, and O.~Kramer, ``Fast and simple
  gradient-based optimization for semi-supervised support vector machines,''
  \emph{Neurocomputing}, vol. 123, pp. 23--32, 2014.

\bibitem{Zemel2000}
R.~Zemel and T.~Pitassi, ``A gradient-based boosting algorithm for regression
  problems,'' \emph{Advances in neural information processing systems},
  vol.~13, 2000.

\bibitem{Loshchilov2016}
I.~Loshchilov and F.~Hutter, ``Sgdr: Stochastic gradient descent with warm
  restarts,'' \emph{arXiv preprint arXiv:1608.03983}, 2016.

\bibitem{Pan2020}
H.~Pan, X.~Niu, R.~Li, Y.~Dou, and H.~Jiang, ``Annealed gradient descent for
  deep learning,'' \emph{Neurocomputing}, vol. 380, pp. 201--211, 2020.

\bibitem{Ruder2016}
S.~Ruder, ``An overview of gradient descent optimization algorithms,''
  \emph{arXiv preprint arXiv:1609.04747}, 2016.

\bibitem{Dogo2018}
E.~M. Dogo, O.~Afolabi, N.~Nwulu, B.~Twala, and C.~Aigbavboa, ``A comparative
  analysis of gradient descent-based optimization algorithms on convolutional
  neural networks,'' in \emph{2018 international conference on computational
  techniques, electronics and mechanical systems (CTEMS)}.\hskip 1em plus 0.5em
  minus 0.4em\relax IEEE, 2018, pp. 92--99.

\bibitem{Watanabe2018}
T.~Watanabe and H.~Iima, ``Nonlinear optimization method based on stochastic
  gradient descent for fast convergence,'' in \emph{2018 IEEE International
  Conference on Systems, Man, and Cybernetics (SMC)}.\hskip 1em plus 0.5em
  minus 0.4em\relax IEEE, 2018, pp. 4198--4203.

\bibitem{Monderer1996}
D.~Monderer and L.~S. Shapley, ``Potential games,'' \emph{Games and economic
  behavior}, vol.~14, no.~1, pp. 124--143, 1996.

\bibitem{Yamamoto2015}
K.~Yamamoto, ``A comprehensive survey of potential game approaches to wireless
  networks,'' \emph{IEICE Transactions on Communications}, vol.~98, no.~9, pp.
  1804--1823, 2015.

\bibitem{Uhlenbeck1930}
G.~E. Uhlenbeck and L.~S. Ornstein, ``On the theory of the brownian motion,''
  \emph{Physical review}, vol.~36, no.~5, p. 823, 1930.

\bibitem{Epperson2013}
J.~F. Epperson, \emph{An introduction to numerical methods and analysis}.\hskip
  1em plus 0.5em minus 0.4em\relax John Wiley \& Sons, 2013.

\bibitem{Chen2019}
Y.~Chen, Y.~Chi, J.~Fan, and C.~Ma, ``Gradient descent with random
  initialization: Fast global convergence for nonconvex phase retrieval,''
  \emph{Mathematical Programming}, vol. 176, pp. 5--37, 2019.

\bibitem{Schwung2022}
D.~Schwung, S.~Yuwono, A.~Schwung, and S.~X. Ding, ``Plc-informed distributed
  game theoretic learning of energy-optimal production policies,'' \emph{IEEE
  Transactions on Cybernetics}, 2022.

\bibitem{Loppenberg2024}
M.~L{\"o}ppenberg, H.~Klopries, J.~Bartsch, and A.~Schwung, ``Foreign object
  separation in bulk good systems using machine learning and image
  processing,'' in \emph{2024 IEEE International Conference on Industrial
  Technology (ICIT)}.\hskip 1em plus 0.5em minus 0.4em\relax IEEE, 2024, pp.
  1--6.

\bibitem{Bergstra2013}
J.~Bergstra, D.~Yamins, D.~D. Cox \emph{et~al.}, ``Hyperopt: A python library
  for optimizing the hyperparameters of machine learning algorithms.''
  \emph{SciPy}, vol.~13, p.~20, 2013.

\bibitem{Arend2022}
D.~Arend, S.~Yuwono, M.~R. Diprasetya, and A.~Schwung, ``Mlpro 1.0-standardized
  reinforcement learning and game theory in python,'' \emph{Machine Learning
  with Applications}, vol.~9, p. 100341, 2022.

\bibitem{Yuwono2023c}
S.~Yuwono, M.~L{\"o}ppenberg, D.~Arend, M.~R. Diprasetya, and A.~Schwung,
  ``Mlpro-mpps-a versatile and configurable production systems simulator in
  python,'' in \emph{2023 IEEE 2nd Industrial Electronics Society Annual
  On-Line Conference (ONCON)}.\hskip 1em plus 0.5em minus 0.4em\relax IEEE,
  2023, pp. 1--6.

\end{thebibliography}

\end{document}